\documentclass[12pt]{article}

\usepackage[preprint]{corl_2024} % Use this for the initial submission.
\usepackage{setspace}
\usepackage{times}
\usepackage{gensymb}
\usepackage{amssymb}
\usepackage{amsmath}
\usepackage{mathrsfs}
\usepackage{bbm}
\usepackage{graphicx}
\usepackage{booktabs}
\usepackage{hyperref}
\usepackage{multirow}
\usepackage{caption}
\usepackage{subcaption}

\title{Transformers for Image-Goal Navigation}

\author{
  Nikhilanj V. Pelluri\\
  Minnesota Robotics Institute, Department of Computer Science and Engineering\\
  University of Minnesota - Twin Cities\\
  United States\\
  \texttt{pellu003@umn.edu} \\
}

\begin{document}
\maketitle

%===============================================================================

\begin{abstract}
    Visual perception and navigation have emerged as major focus areas in the field of embodied artificial intelligence. We consider the task of image-goal navigation, where an agent is tasked to navigate to a goal specified by an image, relying only on images from an onboard camera. This task is particularly challenging since it demands robust scene understanding, goal-oriented planning and long-horizon navigation. Most existing approaches typically learn navigation policies reliant on recurrent neural networks trained via online reinforcement learning. However, training such policies requires substantial computational resources and time, and performance of these models is not reliable on long-horizon navigation. In this work, we present a generative Transformer based model that jointly models image goals, camera observations and the robot's past actions to predict future actions. We use state-of-the-art perception models and navigation policies to learn robust goal conditioned policies without the need for real-time interaction with the environment. Our model demonstrates capability in capturing and associating visual information across long time horizons, helping in effective navigation.

\end{abstract}

\keywords{Visual Navigation, Transformers, Behavior Cloning} 

%===============================================================================

\section{Introduction}
\label{sec:intro}
    Autonomous navigation in environments is a critical capability for modern mobile robots, and has been extensively studied over several decades. Classical approaches to navigation rely on constructing detailed maps of the environment and accurate localization of the robot within the map \cite{thrun2002probabilistic, durrant2006simultaneous, fuentes2015visual}. However, with increasing demand for deploying robots in novel uncontrolled environments such as households, last-mile delivery, etc., constructing accurate and fine-grained maps frequently is often impractical. Robots must now be able to navigate without maps, which means efficient navigation policies require accurate semantic understanding of the scene, efficient exploration and episodic memory, and long-horizon planning with limited knowledge of the environment. Advances in scene understanding and a have led to semantic navigation tasks such as image-goal navigation   \cite{zhu2017target, krantz2023navigating}, object-goal navigation \cite{anderson2018evaluation, chaplot2020object}, etc. receiving significant focus in recent years. \\
    In this work, we consider the specific task of image-goal navigation where robot's navigation objective is specified by an RGB image. We motivate the task with the following scenario: Consider a mobile household robot equipped with an onboard camera tasked with picking up a novel unseen object (say, a new shirt). Since the robot has no prior knowledge about the novel object, it would need other semantic information to understand the object -- an image of the object would serve this purpose effectively. However, in order to navigate efficiently using just this image, the robot would need to accurately understand the semantic information in the image, compare it against its camera observations and navigate towards the goal, while maintain a history of its previous observations and actions.
    \begin{figure}[ht]
        \centering
        \includegraphics[width=1\linewidth]{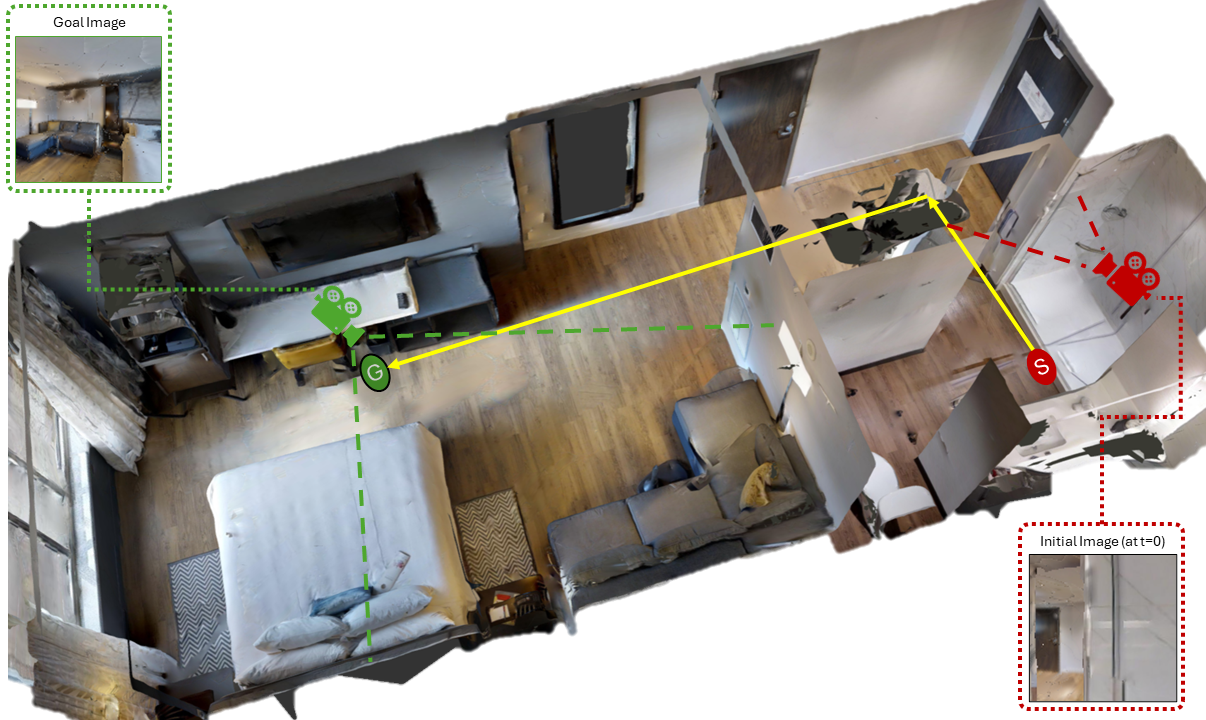}
        \caption{The Image-Goal navigation task. The robot receives a goal image (shown in the green dotted box), and must navigate to find the goal image using only visual observations. In the figure above, the red camera marks the initial viewpoint, with ``S" marking the robot's initial position. The robot navigates to the point marked ``G" described by the goal image, i.e., the view from the green camera. The dashed lines out of the camera indicate the camera's approximate field of view. The yellow arrows indicate one possible path from ``S" to ``G". The figure above shows the ``Pablo" scene from the Gibson training dataset, obtained from \cite{armeni20193d}.}
        \label{fig:ign}
    \end{figure}\\
    Advances in learning based methods for perception and planning have led to the popularity of learnt end-to-end models that takes images as inputs and directly predict actions. These models generally consists of a visual encoder stage to convert input images into features, while a recurrent neural network (RNN) stage maintains history of these features and predicts actions \cite{mirowski2016learning, kulhanek2019vision, yadav2023offline}. However, most RNN based implementations compress history into a single state vector, leading to degraded performance in capturing long-range dependencies \cite{pascanu2013difficulty}. Furthermore, most end-to-end models learn navigation policies using online reinforcement learning and suffer from sample inefficiency - training usually requires extensive interactions in the environment, often in the order of hundreds of millions of episodes. Hence, training these models requires significant compute and time, and is sensitive to hyperparameter tuning \cite{pmlr-v202-eimer23a}.\\
    In recent years, Transformer based models have significantly outperformed RNN based models on sequential modeling tasks, particularly in tasks demanding long horizon understanding. Furthermore, the flexible nature of Transformers enables it to jointly model different modalities such as images, text, actions, etc., which unlocks a powerful capability to correlate input observations and actions \cite{jaegle2021perceiver, driess2023palm}. Propelled by advances in language generation, Transformer models have shown an excellent capability to autoregressively generate very long sequences of tokens \cite{gpt2, Dai2019TransformerXLAL, Yan2021VideoGPTVG}. These models are also easier to train and are able to scale learning with more data.\\
    In our current work, we leverage the effectiveness of Transformer based models to learn policies for image-goal navigation. We jointly model observations, goals and the robot's previous actions to learn goal-conditioned navigation policies, and generate arbitrarily long sequences of actions to reach image goals.\\
    We use state-of-the-art models in image-goal navigation to create datasets of expert trajectories to solve the task, and then mimic these expert policies using a behavior cloning approach. This approach circumvents the need to access an environment online and enables us to train a model with significantly lower compute and time requirements.\\
    % We further explore a local planning approach using only observation and goal images. Our approach does not require depth or position sensors. However, in practice, we find adding depth improves accuracy.
%===============================================================================

\section{Related Work}
\label{sec:Related Work}

\subsection{Image-Goal Navigation}
Visual Navigation, i.e., robot navigation using visual input has been extensively studied in literature \cite{desouza2002vision}. We particularly focus on image-goal navigation in our work. Several different approaches have been suggested to solve this task, but they can be primarily categorized into:\\
\begin{itemize}
    \item \textbf{Map Based Approaches:} These approaches focus on first generating a full map of the environment, and using the map for localizing the agent and planning a path to the goal. While some works focus on building accurate metric maps with embedded semantic information \cite{occupancy, chaplot2020object, huang2023visual, kwon2023renderable} of the environment, some other works focus on learning a topographical map instead, and use graph-based planning methods for global navigation \cite{chaplot2020neural, hahn2021no, kim2023topological}. However, map-based approaches are heavily reliant on accurate mapping and precise localization of the robot, which may not be feasible, especially in dynamic scenarios.
    \item \textbf{Learning Based Approaches:} The development of high-fidelity physics simulators such as Habitat \cite{savva2019habitat}, AI2-THOR \cite{ai2thor} and photorealistic scene datasets such as Gibson \cite{gibson}, Matterport-3D \cite{ramakrishnan2021hm3d}, etc. has enabled agents to be trained in simulation using deep reinforcement learning techniques. Mishkin et al. \cite{Mishkin2019BenchmarkingCA} showed that while classical mapping and localization based approaches worked well with depth information, learned policies performed better with RGB-only input.  \cite{zhu2017target, zer, mezghan2022memory, yadav2023offline, yadav2023ovrl, bono2024endtoend} all utilize the similar architecture of a visual encoder for image representation and a recurrent network for episodic memory.
\end{itemize}
Our approach falls under the learning based approach. We do not generate an explicit map of the environment and instead rely on the learned policy to guide the agent towards the goal. However, as detailed earlier in the Section \ref{sec:intro}, recurrent networks do not perform well with long horizon tasks. We believe new state-of-the-art sequential architectures, particularly Transformer models hold the key to efficient long horizon image-goal navigation.

\subsection{Transformers}
Sequence models have seen rapid progress in recent times, particularly since the introduction of the Transformer model \cite{vaswani2017attention}. Transformer based models have shown remarkable ability to model several different types of inputs including vision \cite{vit, dinov2}, vision and language \cite{alayrac2022flamingo}, multi-modal inputs including actions \cite{10123038, shridhar2022peract}, etc. This versatility of Transformer is useful to model observations and actions together, which could potentially help build better ``world knowledge" in the model. \\
Some works such as \cite{scenemem} use the Transformer architecture for visual navigation, but do not specifically target the image-goal or object-goal navigation tasks. Several works such as \cite{pashevich2021episodic, chen2021hamt} apply Transformers to visual-language navigation. However, these works generally learn navigation using language instructions, whereas we focus on goal-oriented vision-only navigation.\\
In this work, we primarily draw from the work of Decision Transformer \cite{chen_decision_2021}, Trajectory Transformer \cite{trajectory} and PACT \cite{bonatti2023pact}. We adapt the formulation of the Decision Transformer model to fit our image-goal navigation task by conditioning the model to pay attention to the goal. We achieve this by inserting goal tokens into the input tokens sequence. We use a similar decoder-only architecture that autoregressively generates action tokens.\\
Training with online reinforcement learning (RL) requires access to rewards from a simulated environment over hundreds of millions of episodes, which demands humongous compute resources and time. Furthermore, it has been shown that Transformer models are considerably difficult to train in an online RL setting \cite{parisotto2020stabilizing}.\\
To circumvent this, we adopt a goal-conditioned behavior cloning (GCBC) approach. This allows us to learn scalably using an expert agent's experience.

\subsection{Behavior Cloning}
Behavior Cloning (BC) has been used to learn several kinds of robotic tasks, particularly where online RL techniques cannot be applied or is too expensive, such as autonomous driving \cite{pomerleau1988alvinn}, real robot manipulation \cite{jang2021bc}, social navigation \cite{karnan2022scand}, etc. Specifically, we follow an approach similar to \cite{ding2019goal, lynch2019play} to learn ``goal-conditioned" policies from behavior cloning. \\
Most BC works rely on large datasets of inputs and corresponding actions taken by human experts \cite{mandlekar2018roboturk}. However, collecting such a large number of human navigation trajectories entail significant effort and time. While the Habitat-Web dataset \cite{rramrakhya2022}, with 77k human demonstrations, has been created for the object-goal navigation task, we are not aware of a similar dataset for image-goal navigation.\\
Hence, we rely on an expert agent -- an agent trained on episodes in the training set, to generate training data for BC. We collect all trajectories from this pre-trained agent, filter out unsuccessful episodes, and create an expert dataset containing trajectories that reach the image goals for the episodes. RLBench \cite{james2020rlbench} generates expert trajectory data for robot manipulation using hand-engineered motion planners, whereas we use a learned policy to generate new trajectory data.\\
We train our decoder model to mimic expert trajectories, closely related to the approach taken by \cite{shah2023vint, rt_1}.

\section{Methods}
\label{sec:result}
\subsection{Problem Statement}
\subsubsection{Image-Goal Navigation}
We consider the task of image-goal navigation. An agent is initialized at a random pose $T_0$ in an unseen environment, and is tasked with reaching a goal pose $T_G$, described by the agent's observation at the goal pose, i.e., RGB image $I_G$. The agent take an action $a_t$ in the environment using egocentric observations $o_t$ at every timestep $t$, and receives a reward $r_t$, including a bonus reward for finishing within a radius $d_g$ of the goal.\\
The problem can be formalized as a Partially Observable Markov Decision Process (POMDP) defined by the tuple $<\mathcal{S}, \mathcal{A}, \mathcal{T}, \mathcal{R}, \mathcal{O}>$  with states and goals ${s,g} \in \mathcal{S}$, actions $a \in \mathcal{A}$, transition dynamics $p(s'|a,s) \in \mathcal{T}$, rewards $r(s,a,g) \in \mathcal{R}$, sensor observations $o \in \mathcal{O}$. It is worth noting here that in our task, states and goals belong to the same space, since any state can be considered a goal.\\
Solving the task consists of learning a policy $\pi(a|o,G)$ that maximizes the cumulative return over an episode.
\[
J(\pi) = \mathbb{E}_{g \sim p_g \mathcal{S}, \tau \sim d_{\pi}(.|g)}[\sum_{t} \gamma^t r(s_t, a_t, g)]
\]

\subsubsection{Goal-Conditioned Supervised Learning/Behavior Cloning}
We follow the formulation of \cite{ghosh2021learning} for Goal-Conditioned Supervised Learning (GCSL), also known as Goal-Conditioned Behavior Cloning (GCBC).\\
We assume access to a dataset $D^*$ of demonstrations $\{\tau_0, \tau_1....\tau_N\}$ from an expert policy $\pi^*$. Each demonstration $\tau_i$ contains a goal image $I_{G}$ and a sequence of observations $o_t$, actions $o_t$ taken by the expert policy, and the corresponding rewards received $r_t$. 
\[
\tau_i = [I_{G};\{o_t,a_t,r_t\}_{t=1}^T]_{i=0}^N
\]
The objective of GCBC is to learn a policy $\pi_\theta(a|o,I_G)$ by minimizing the error between the output actions and the expert's actions, such that
\[
\pi_{\theta} = \arg\min_{\theta} \sum_{\tau \in D^*} \sum_{\{o,I_G\} \in \tau} L(\pi(a|o,I_G), \pi^*(a|{o,I_G}))
\]

% We follow the formulation of \cite{yadav2023ovrl} and define the reward at each timestep $t$ as \[
% r_t = (d_{t-1} - d_t) - \gamma + \mathbbm{1}(d_t < R_g \ \& \ a_t = \texttt{STOP})
% \]
%,where $d_t$ is the geodesic distance to the goal, $\gamma$ is a time penalty to efficient navigation, and $R_g$ is defined as the goal radius.
% The problem can be formalized as a goal-conditioned Partially Observable Markov Decision Process (POMDP) defined by a tuple $<\mathcal{S}, \mathcal{A}, \mathcal{T}, \mathcal{T}, \mathcal{R}>$
% Solving the task consists of learning a policy $\pi(a|o,g):\mathcal{O}, \mathcal{G} \rightarrow \mathcal{A}$ to maximize the cumulative reward over an episode. 

% \subsubsection{Rewards and Success Criteria}
% We follow the reward setup from \cite{yadav2023ovrl}, wherein the agent receives a proportional positive reward for moving towards the goal, and a bonus reward of $+2.5$ for reaching within a radius of 1 meter to the goal.\\

\subsection{Model}
\begin{figure}[htbp]
    \centering
    \includegraphics[width=0.75\linewidth]{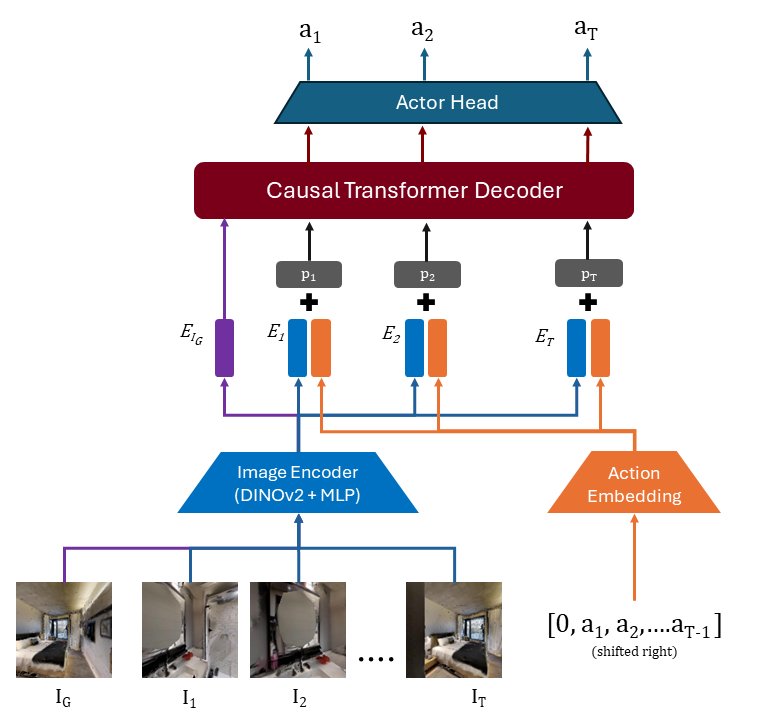}
    \caption{The full architecture of our model. The goal image $I_G$ and observation images $I_1, I_2,...I_T$ are embedded into tokens $E_{I_G}$ and $E_{I_1}, E_{I_2},....E_{I_T}$ respectively using a pre-trained DINOv2 model, actions $a_1, a_2,...a_T$ are passed to a specific lookup embedding. To enable autoregressive generation, the input actions are shifted right by one step. Image and action embeddings are interleaved to form input embeddings $E_1, E_2,...E_T$. Learnable position embeddings $p_1, p_2,...p_T$are added before passing all inputs to a Transformer decoder with a causal attention mask. The decoder outputs logits, corresponding to the most likely action. An actor head samples the actions from these logits.}
    \label{fig:decoder}
\end{figure}

Our model is built as a decoder-only Transformer model, drawing from the works of GPT \cite{gpt1}, Decision Transformer (DT) \cite{chen_decision_2021}, ViNT \cite{shah2023vint}, RT-1 \cite{rt_1}, etc.\\
The model takes as input the goal image ($I_G$), a sequence of observation images (${I_i....T_{i+T}}$, $T$ being the sequence length) and the actions taken by the expert (${a_i....a_{i+T}}$).
% and the corresponding rewards received by the expert (${r_i....r_{i+T}}$).
Notably, unlike DT, we do not condition the model on the returns-to-go and instead condition the model's actions on the goal token.\\
Our model has 27M trainable parameters, excluding the frozen parameters of the image encoder model.\\
The full architecture is shown in Figure \ref{fig:decoder}.

\subsubsection{Input Representation}
\textbf{Image Encoding:} We first encode the goal image $I_G$ and the observation images $I_t$ to corresponding embeddings $E_{I_i} \in \mathbb{R}^d $ using an image encoder. We use the ``ViT-B/14" version of the DINOv2 model \cite{dinov2} with registers \cite{dinoreg} as the encoder, which reports robust performance on a variety of downstream vision tasks. The DINOv2 model is based on the Vision Transformer architecture \cite{vit}, which splits images into patches (here, of size 14x14), before passing linear projections of the patches to a Transformer encoder consisting of multiple layers of multi-head self-attention (MHSA) blocks. We freeze the pre-trained weights for the model and do not train the feature extraction layers of the model.\\
To improve learning from finer features, we concatenate the patch tokens from the final four layers of the DINOv2 model, and along with the output of the \texttt{[CLS]} token, which contains global context about the image. The concatenated feature vector is passed to a 2-layer multi-layer perceptron module (MLP) with an output dimension $d$.

\textbf{Action Embedding:} The categorical input actions are encoded into $d$-dimensional embeddings $E_{a_i}  \in \mathbb{R}^d $ using a lookup table with learnable weights. 
% We use the \texttt{nn.Embedding} layer in PyTorch \cite{pytorch} for this purpose.\\
It is important to note that during training, the actions are shifted to the right, to condition the model for autoregressive action generation. \\
% \textbf{Reward Embedding:} The rewards are input as floating-point values, which are encoded to $d$-dimensional embeddings $E_{r_i}  \in \mathbb{R}^d $, using a single linear layer.\\
Following \cite{chen_decision_2021}, we first interleave the input observation and action sequences, such that the combined input sequence becomes 
\[\mathbf{e} = [E_{I_1}, E_{a_1}, E_{I_2}, E_{a_2}, ....]\]

\textbf{Position Embedding:} We add a learnable position embedding $p_i \in \mathbb{R}^{d} $ to each element sequence. We add the same position embedding to tokens of each timestep, i.e., the same embedding $p_i$ is added to $E_{I_i}$ and $E_{a_i}$.Hence, the position embedding $\mathbf{p} \in \mathbb{R}^{2T \times d} = [p_1, p_1, p_2, p_2,...]$\\
Since we want to learn policies conditioned on the goal image $I_G$, we insert the goal image embedding $E_{I_G}$ at the beginning of the sequence. This is similar to the approach taken by \cite{ghosh2021learning} and \cite{trajectory}. Inserting the goal embedding at the beginning enables all embeddings in the sequence to attend to the goal, since we use a causal attention mask in the decoder.\\
The full input to the Transformer is 
\[
\mathbf{x} = [E_{I_G}; \mathbf{e} + \mathbf{p}]
\]
\subsubsection{Transformer Decoder}
Our decoder-only Transformer sequence model consists of $L = 12$ layers, with each layer composed of $H = 8$ multi-head self-attention (MHSA) blocks and feed-forward network dimension $d_{FFN} = 1024$. Following \cite{preln}, the LayerNorm (LN) operation is applied before each block. We apply a causal self-attention mask, which prevents the model from ``looking ahead" and copying future actions. The decoder accepts embeddings of dimension $d = 384$ as input.\\
Scaled dot-product attention operation of a single head $h$ for an input token sequence $\mathbf{x} \in \mathbb{R}^{T \times d}$ is defined as:
\begin{gather*}
Q = \mathbf{x}{W_h}^Q  ;  K = \mathbf{x}{W_h}^K  ;  V = \mathbf{x}{W_h}^V\\
\text{where} \ \{{W_h}^Q, {W_h}^K, {W_h}^V\} \in \mathbb{R}^{d \times d}\\
\textsc{Attn}_{h}(Q, K, V) = \textit{softmax}(\frac{Q \cdot K^T}{\sqrt{d_k}}) \cdot V
\end{gather*}

The attention values from each head are concatenated together to form the MHSA output:
\begin{gather*}
\textsc{MHSA}(\mathbf{x}) = \textsc{Concat}[\textsc{Attn}_{1}, \textsc{Attn}_{2}.....\textsc{Attn}_{H}]W^O\\
\text{where} \  W^O \in \mathbb{R}^{Hd \times d}
\end{gather*}

The computations in a single layer $l$ of the decoder for the input token sequence $\mathbf{x}$ can be represented as:
\begin{gather*}
\mathbf{y}_{l} = \textsc{MHSA}(\textsc{LN}(\mathbf{x}_{l})) + \mathbf{x}_{l}\\
\mathbf{z}_{l} = \textsc{FFN}(\textsc{LN}(\mathbf{y}_{l})) + \mathbf{y}_{l}
\end{gather*}
\subsubsection{Actor model}
The actor model has a multi-layer perceptron (MLP) network that maps the $d$-dimensional output from the decoder to logits of the actions space.\\
During training, we apply a softmax operation over the logits, and the highest probability tokens are chosen as the output actions.\\
However, during autoregressive action generation in \ref{sec:autoreg}, we find empirically that greedily picking the highest probability token leads to degenerate scenarios such as endlessly repeating the same token or the agent getting stuck in closed spaces. To resolve this, we apply a top-$k$ ($k = 2$) sampling strategy to pick the next action. Sampling actions improves the exploration capability of the agent, while slightly increasing the risk of higher distribution shift away from the expert policy.

\section{Experimental Results and Analysis}
\label{sec:results}
\subsection{Environment Setup}
We use \texttt{v0.2.1} of the Habitat simulator \cite{savva2019habitat}.\\
Following \cite{yadav2023ovrl}, our agent contains only an egocentric camera sensor with a field-of-view of 90\degree, which returns observations as RGB images of size 128 pixels x 128 pixels. We do not use depth observations or GPS+Compass information.\\
The action space of the agent is composed of 4 discrete actions: [\texttt{STOP}, \texttt{MOVE\_FORWARD}, \texttt{TURN\_LEFT}, \texttt{TURN\_RIGHT}]. The agent moves forward by 0.25m on the \texttt{MOVE\_FORWARD} action, and turns 30\degree on \texttt{TURN\_LEFT} or \texttt{TURN\_RIGHT}.

\subsection{Dataset details}
\label{section:dataset}
We run the agent in scenes from the Gibson dataset for Habitat-Sim \cite{gibson}. We use the training split from \cite{zer}, which consists of 9000 episodes sampled from each of the 72 scenes in the Gibson training set. The test split from \cite{hahn2021no} is used for evaluation.  In both training and test splits, the episodes are split into ``straight" and ``curved" episodes. We refer to \cite{hahn2021no} for the categorization criteria: ``in ‘straight’ episodes, the ratio of shortest path geodesic-distance to euclidean-distance between the start and goal locations is $<$1.2 meters and rotational difference between the orientation of the start position and goal image is $<$45\degree. All other start-goal location pairs are labeled as curved episodes."\\
Episodes in each category are further categorized depending on the geodesic distance between start and goal locations into easy (1.5m-3m), medium (3m-5m) and hard (5m-10m) difficulties.

\subsection{Evaluation Metrics}
\label{section:evaluation_metrics}
We refer to \cite{anderson2018evaluation} which introduces two standard metrics for goal navigation :
\begin{itemize}
    \item \textbf{Success Rate:} The agent is considered to have successfully completed the episode if it calls the \texttt{`STOP'} action within a radius of 1 meter from the goal position.
    \item \textbf{Success weighted by Path Length (SPL):} This metric measures the ratio of the shortest path versus the path taken by the agent for successful episodes, thereby giving significant weightage to reaching the goal efficiently.
\end{itemize}

\subsection{Collecting expert trajectories for Behavior Cloning}
\label{section:data_collection}
We collect a dataset $D^*$ of expert trajectories using the agent from OVRL-v2 (``OVRL agent").The agent with pre-trained weights (from \cite{wasserman2022lastmile}) gathers trajectories from episodes in the training split. Similarly, we collect the agent's trajectories on the test split for evaluation.

For training and evaluation, we filter out episodes where the OVRL agent was unsuccessful. The trajectories where the OVRL agent successfully navigates to the goal are taken as expert trajectories.

The success rate of the OVRL agent in each category of training and test episodes is given in Table \ref{tab:sling_accuracy}.

\begin{table}[h!]
    \centering
    % \begin{tabular}{|c|c|c|c|c|}
    % \hline
    %     Split & Path Type & Easy  & Medium & Hard \\
    % \hline
    %     Train \cite{zer} & Curved & 117755 (87\%)  & 176525 (81.7\%) & 172133 (80.8\%) \\
    % \hline
    %     Test \cite{hahn2021no} & Curved & 475 (47.5\%)  & 441 (44.1\%) & 319 (31.9\%) \\
    % \hline
    %     Test \cite{hahn2021no} & Straight & 584 (58.4\%)  & 495 (49.5\%) & 323(40.1\%\text{*}) \\
    % \hline
    % \end{tabular}
    \begin{tabular}{@{}lccccc@{}}
        \toprule
        Split & Path Type & \multicolumn{3}{c}{Number of episodes (Success\% of total)} \\
        \cmidrule{3-5}
        & & Easy & Medium & Hard \\
        \midrule
        Train \cite{zer} & Curved & 117755 (87.0\%) & 176525 (81.7\%) & 172133 (80.8\%) \\
        Test \cite{hahn2021no} & Curved & 475 (47.5\%) & 441 (44.1\%) & 319 (31.9\%) \\
        Test \cite{hahn2021no} & Straight & 584 (58.4\%) & 495 (49.5\%) & 323 (40.1\%\text{*}) \\
        \bottomrule
    \end{tabular}
    \caption{The number of successful episodes and their proportion out of all episodes, collected by the pre-trained agent from OVRL-v2 \cite{yadav2023ovrl} on different categories of episodes in the Gibson dataset. \text{*}From 805 total collected episodes for this split.}
    \label{tab:sling_accuracy}
\end{table}

We were unable to access the OVRL agent's actions during the local navigation stage of \cite{wasserman2022lastmile}, since it uses a local geometric planner disconnected from the global navigation stage. Hence, the final observation and action in the expert dataset, i.e., $o_T$ may, in a few cases, not correspond exactly to the goal image $I_G$.

\subsection{Training}
We train the model on 96K expert trajectories randomly sampled from a mix of ``easy", ``medium" and ``hard" difficulty trajectories from the ``curved" split. For each trajectory during training, we extract a contiguous subsequence of length $T=8$, with a randomly sampled start position.\\
The model is trained using the standard categorical cross-entropy loss. We use the AdamW optimizer with a weight decay of 0.01 and a custom epoch-wise decaying learning rate schedule, starting at 1e-4.\\
We train the model for 20 epochs on an NVIDIA 1080Ti GPU with a batch size of 16.

\subsection{Evaluation}
We evaluate the model's performance on two questions: (1) how accurately did our model learn the expert policy [\ref{sec:supervised}] ? and (2) how does the learned policy perform in an online setting, i.e., with autoregressive action generation [\ref{sec:autoreg}] ?

\subsubsection{Learning the expert policy}
\label{sec:supervised}
We measure the accuracy between the actions taken by the expert and the actions taken by our learned agent, given the same goals and observations. We first run inference over non-overlapping sliding windows on expert trajectories from the test set, and measure the accuracy cumulatively across all time windows for a trajectory.  The accuracy on various test splits is reported in Table \ref{tab:offline_accuracy_curved}.

\begin{figure}[h]
    \centering
    \begin{subfigure}{\linewidth}
        \centering
        \includegraphics[height=2.1in]{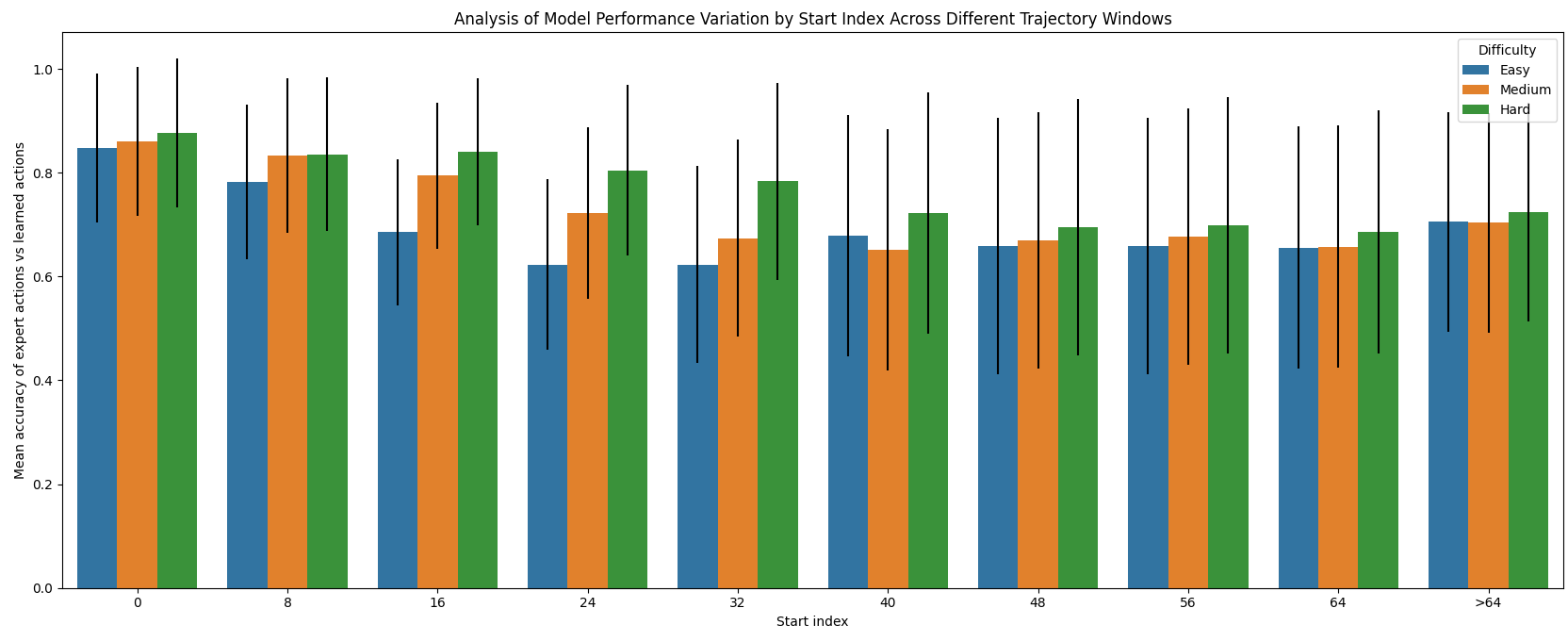}
        \caption{Performance on curved episodes.}
        \label{fig:window_accuracy_curved}
    \end{subfigure}
    \centering
    \begin{subfigure}{\linewidth}
        \centering
        \includegraphics[height=2.45in]{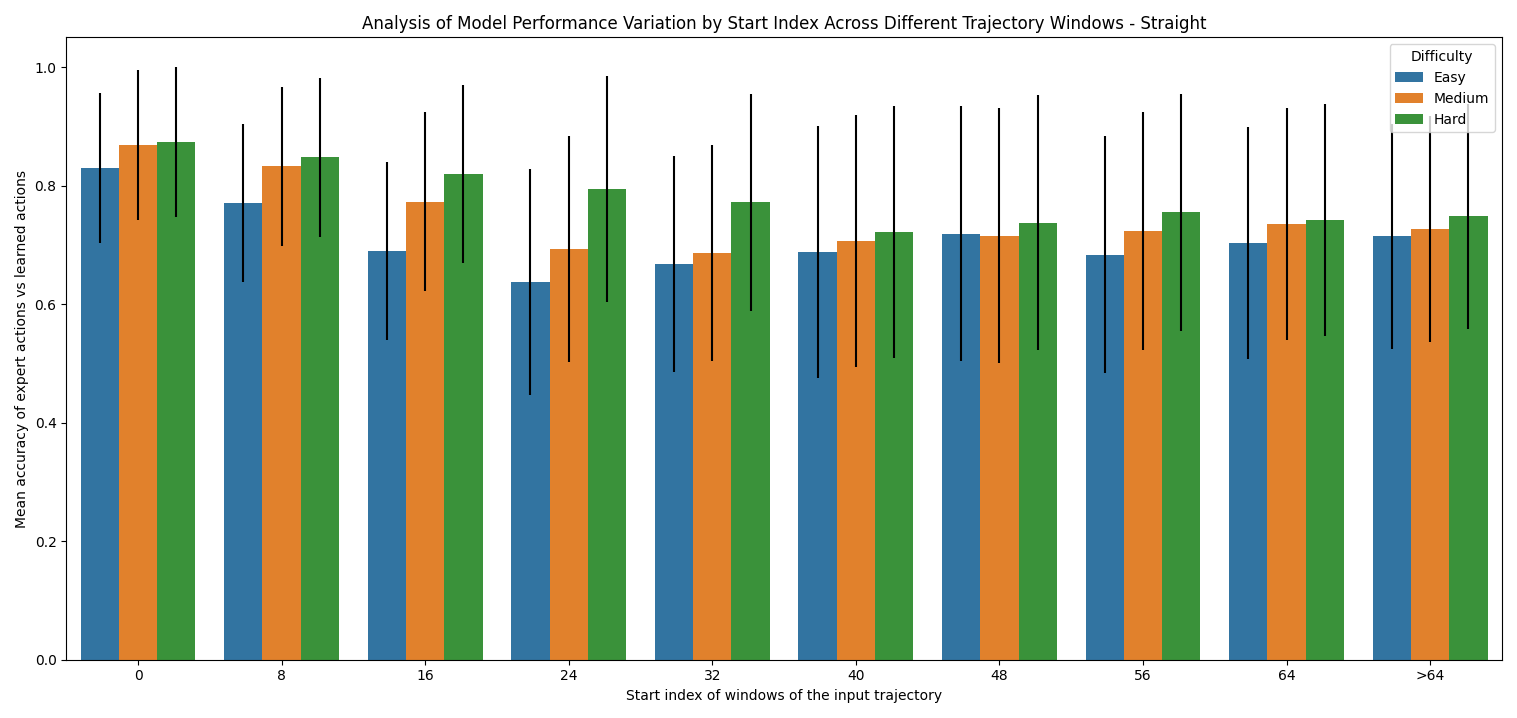}
        \caption{Performance on straight episodes.}
        \label{fig:window_accuracy_straight}
    \end{subfigure}
    \caption{Model accuracy at different start indices of non-overlapping windows of input trajectory, for curved (\ref{fig:window_accuracy_curved}) and straight episodes (\ref{fig:window_accuracy_straight}). Results indicate model accuracy is higher at the episode's beginning and lower near its middle.}
    \label{fig:window_accuracy}
\end{figure}

\begin{table}[hb]
    \centering
    % \begin{tabular}{|c|c|c|c|c|}
    % \hline
    %      Test split & Easy & Medium & Hard\\
    % \hline
    %      Curved & 75.45\% & 77.98\% & 79.31\%\\
    % \hline
    %      Straight & 77.44\% & 79.83\% & 80.49\% \\
    % \hline
    % \end{tabular}
    \begin{tabular}{@{}lccc@{}}
        \toprule
        Test Split & Easy & Medium & Hard \\
        \midrule
        Curved & 75.45\% & 77.98\% & 79.31\% \\
        Straight & 77.44\% & 79.83\% & 80.49\% \\
        \bottomrule
    \end{tabular}
    \caption{Accuracy of our model's actions vs expert actions at various difficulty levels of test episodes from the Gibson dataset, when evaluated over non-overlapping windows. We were unable to collect all trajectories on the ``Straight-Hard" split.}
    \label{tab:offline_accuracy_curved}
\end{table}
We further analyze this question: \textit{does the model's performance differ as the distance between goal and observations changes, i.e., as move further along the trajectory ?} To answer this question, we measured the accuracy separately in each non-overlapping window, and analyzed it against the start index. The results are shown in Figure \ref{fig:window_accuracy}.\\

\begin{figure}
    \centering
    \includegraphics[width=0.8\linewidth]{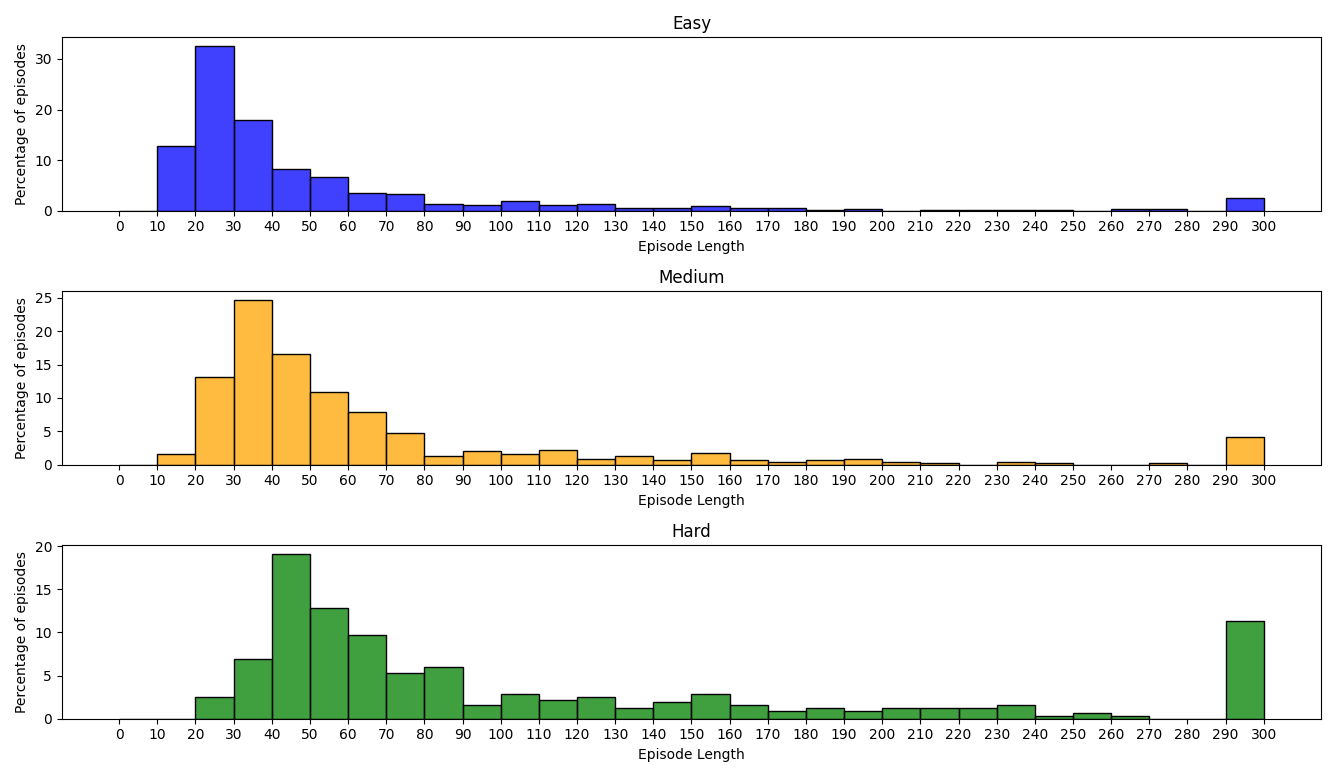}
    \caption{Distribution of episode lengths across various difficulty splits from the ``Gibson-curved" training set.``Easy" episodes are considerably shorter than ``Medium" and ``Hard" episodes.}
    \label{fig:traj_lengths}
\end{figure}

We see that the average accuracy is higher in the initial windows, and decreases marginally at the middle of the trajectory. This could be attributed to the distribution of the input trajectory data, since the expert trajectories are biased to be shorter. This behavior is particularly pronounced in the easy trajectories, which have a shorter episode length. We also observe that the variance in accuracy at the middle of the trajectory is higher than at the start. We show the distribution of expert trajectory lengths across various difficulty splits from the ``Gibson-curved" training set in figure \ref{fig:traj_lengths}.\\

A perplexing question that arises from table \ref{tab:offline_accuracy_curved} and figure \ref{fig:window_accuracy} is that the accuracy on the ``Hard" difficulty trajectories is higher than the ``Easy" and ``Medium" trajectories. We posit that this could be due to the learned policy having ``learnt" the exploratory policy of the expert -- ``Hard" trajectories require considerably more exploration than the others, and exploration strategy is generally similar across different episodes. A similar observation is made with ``Medium" trajectories, where the accuracy is higher than ``Easy" trajectories.
\\

\subsubsection{Autoregressive action generation}
\label{sec:autoreg}
We evaluate the action generation capability of our learned agent on unseen environments from the Gibson environment, following the same test split from \cite{hahn2021no} as detailed in section \ref{section:dataset}. We refer to section \ref{section:evaluation_metrics} for details on the success criteria.\\
The agent is initialized in the environment with the goal image. The agent maintains a ``first-in-first-out" (FIFO) buffer of observations and actions, with a buffer size equal to the decoder model's context length ($T=8$).

We evaluate our model against the following baselines, with results and pre-trained weights provided by \cite{wasserman2022lastmile}:
\begin{enumerate}
    \item \textbf{Behavior Cloning with Spatial Memory:} Observation and goal images are encoded using a ResNet-18 \cite{resnet} model, while the depth $D_t$ is used to construct a spatial metric map, and navigation learnt using behavior cloning (BC).
    \item \textbf{Behavior Cloning with GRU:} Similar to the above model, observations and goal are encoded using the ResNet-18 model, while a Gated Recurrent Unit \cite{gru} network is used to maintain state. BC is used to train the navigation policy.
    \item \textbf{OVRL-v2} \cite{yadav2023ovrl}: This model uses a Vision Transformer \cite{vit} model to encode observation and goal images, while an LSTM-based \cite{lstm} neural network provides state memory. Navigation policy is trained using policy gradient methods over 500M episodes. We use this model to generate expert trajectories data for our model.\\
    We were able to independently verify the results provided by \cite{wasserman2022lastmile} for this model.
\end{enumerate}

The results of our evaluation are shown in table \ref{tab:autoreg_results}. We observe that our behavior cloned model outperforms other behavior cloned models on most data splits. We further note that we use an input image size of 128x128, which is significantly smaller than other BC models, which use an input image size of 480x640 \cite{hahn2021no}.

\begin{table}[ht]
    \centering
    \begin{tabular}{@{}lcccccccc@{}}
        \toprule
        \multirow{2}{*}{Method} & \multicolumn{2}{c}{Overall} & \multicolumn{2}{c}{Easy} & \multicolumn{2}{c}{Medium} & \multicolumn{2}{c}{Hard} \\
        \cmidrule(lr){2-3} \cmidrule(lr){4-5} \cmidrule(lr){6-7} \cmidrule(lr){8-9}
        & Success$^\uparrow$ & SPL$^\uparrow$ & Success$^\uparrow$ & SPL$^\uparrow$ & Success$^\uparrow$ & SPL$^\uparrow$ & Success$^\uparrow$ & SPL$^\uparrow$ \\
        \midrule
        \multicolumn{9}{c}{Dataset split: Gibson-curved}\\
        \midrule
        Behavior Cloning w/ Spatial Memory\textsuperscript{\textdagger} &
         1.30 & 1.10 & 3.10 & 2.50 & 0.80 & 0.70 & 0.20 & 0.10 \\
        Behavior Cloning w/ GRU State\textsuperscript{\textdagger} & 
         1.70 & 1.30 & 3.60 & 2.80 & 1.10 & 0.90 & 0.50 & 0.30 \\
         \textbf{Ours}\text{*} & 6.78 & 6.37 & 16.00 & 15.08 & 3.67 & 3.49 & 0.67 & 0.54 \\
        OVRL-v2\textsuperscript{\textdagger} (Expert) & 45.60 & 28.00 & 53.60 & 31.70 & 47.60 & 30.20 & 35.60 & 21.90 \\
        \midrule
        \multicolumn{9}{c}{Dataset split: Gibson-straight}\\
        \midrule
        Behavior Cloning w/ Spatial Memory\textsuperscript{\textdagger} & 12.50 & 12.10 & 24.80 & 23.90 & 11.50 & 11.20 & 1.30 & 1.20 \\
        Behavior Cloning w/ GRU State\textsuperscript{\textdagger} & 19.50 & 18.70 & 34.90 & 33.40 & 17.60 & 17.00 & 6.00 & 5.90 \\
        \textbf{Ours}\text{*} & 15.06 & 14.04 & 28.33 & 26.13 & 11.33 & 10.53 & 4.00 & 3.65 \\
        OVRL-v2\textsuperscript{\textdagger} (Expert) & 44.90 & 30.00 & 53.60 & 34.70 & 48.60 & 33.30 & 32.50 & 21.90 \\
        \bottomrule
    \end{tabular}
    \caption{
    Results from evaluating our trained agent on test scenes from the Gibson dataset.  \text{*}Note that we evaluate on a random sample of 300 out of total 1000 episodes, and limit our episode length to 100 for easy episodes and 350 for medium and hard episodes. \textsuperscript{\textdagger}Results from \cite{wasserman2022lastmile}}
    \label{tab:autoreg_results}
\end{table}

\begin{figure}[htbp]
    \centering
    \includegraphics[width=0.5\linewidth]{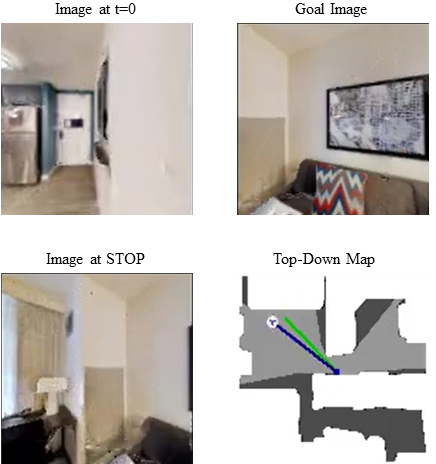}
    \caption{Model's performance on a successful ``easy" difficulty episode from the ``curved" split of the Gibson test set. }
    \label{fig:ep2435}
\end{figure}

\begin{figure}[htbp]
    \centering
    \includegraphics[width=0.5\linewidth]{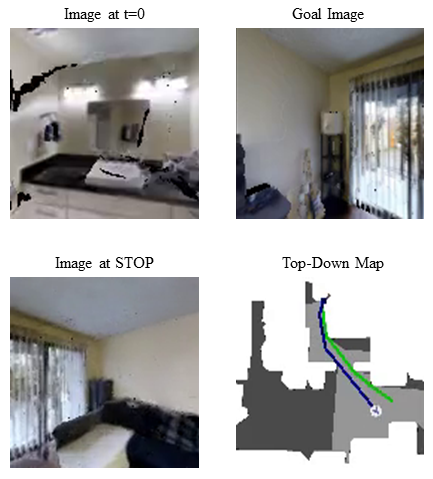}
    \caption{Model's performance on a successful ``medium" difficulty episode from the ``curved" split of the Gibson test set. }
    \label{fig:ep505}
\end{figure}

We show some qualitative results of the model's performance in figures \ref{fig:ep2435}, \ref{fig:ep505}.

\subsection{Failure Modes}
We now analyze some cases where our agent fails:
\begin{itemize}
    \item \textbf{Small/Invisible obstacles:} In several cases, the agent encounters obstacles that are not visible in its onboard camera. Since our model relies exclusively on images from the camera, the model is unable to ``understand" the presence of a obstacle, particularly when the height of the obstacle is less than the height of the camera (for example, a sofa or a coffee table).
    \item \textbf{Inability to recover from collisions:} An extension of the above issue is that since our model does not access rewards from the environment nor other information such as pose or depth, the agent cannot recover from collisions with the environment. For example, the agent may run into a wall partially, but since it has no way of understanding this scenario, the agent continues trying to go straight until it expends all its steps budget. Furthermore, since we use successful expert trajectories, which are biased against collisions, the model does not learn the behavior required to recover from collisions.
    \item \textbf{Dataset specific considerations:} As \cite{yadav2023ovrl} observe in their failure analysis, about 6\% of episodes in the dataset have goal images that are semantically meaningful - such as blank walls. Since we rely exclusively on image feature extraction, not having distinguishable features significantly affects the model's ability to understand the scene. Another mode of failure is when the agent needs to pass through narrow hallways to reach the goal. Accuracy in such cases demands fine-grained control of the robot, which is difficult to learn with a global navigation policy designed for long-horizon navigation.
\end{itemize}

% \pagebreak
\section{Conclusion}
\label{sec:conclusion}
In this work, we presented a decoder-only Transformer model for image-goal navigation. We trained the model using a Goal-conditioned Behavior Cloning (GCBC) approach, thereby avoiding the need for interactions with the environment. Our model significantly outperforms other behavior cloning baselines on the image-goal navigation task. While our GCSL model is easily trained under computational constraints in a relatively short period of time, we find that it underperforms state-of-the-art online RL based methods, which were trained for millions of episodes. This leaves further room across various axes of improvement to the current model. The flexibility of the Transformer architecture allows the same model to be trained with other input modalities such as text, speech, etc.

\section{Future Work}
Our work opens up several interesting areas of further research.\\
In our current work, the Transformer Decoder uses self-attention layers to learn correlations between the goal features, observations and previous actions. However, training such a model end-to-end might result in subpar performance at low training scale since the priors for the model are too weak -- the model has to learn image matching as well as navigation at the same time. We believe that pre-training or auxiliary training on explicit image matching tasks such as wide-baseline stereo matching, optical flow, etc. would help decouple learning in the perception and navigation modules. This could potentially boost performance on the overall task, since the decoder can focus on planning and navigation.\\
Another avenue for improvement is the fact that we used the vanilla Transformer implementation from \cite{vaswani2017attention}, where the self-attention computation grows quadratically with input context length. However, in recent years, several improvements have been suggested to this architecture, particularly towards increasing the context length and reducing computational costs. Using these improved architectures could potentially help navigate efficiently in particularly complex and long-horizon scenarios.\\
Additionally, in this work, we focus on behavior cloning to train our policies quickly and efficiently. However, there remains a significant gap to fill between behavior cloned policies and efficient policies in the real world. Online fine-tuning could help bridge this gap \cite{ramrakhya2023pirlnav}. The gap between the simulated environments and real world scenarios also points to the fact that fine-tuning the model in the real world is an essential step towards robust deployment of these models.
%===============================================================================

\clearpage
% The acknowledgments are automatically included only in the final and preprint versions of the paper.
\acknowledgments{
We thank Houjian Yu and members from CHOICE Lab for their inputs and support. We would also like to thank Justin Wassermann from the University of Illinois Urbana-Champaign for sharing their code and pre-trained weights for the OVRL-v2 model. This work was made possible due to the continued support of the Minnesota Robotics Institute and the Department of Electrical and Computer Engineering at the University of Minnesota.
}

%===============================================================================

\bibliography{references}  % .bib
\clearpage

\appendix

\section{Example Expert Trajectories}
The following shows some example expert trajectories from the test split of Gibson dataset. The trajectories are collected using the pre-trained model from OVRL-v2 \cite{yadav2023ovrl}.\\
Episodes in the dataset are split into ``straight" and ``curved" path types depending on the path, which are further split into ``easy", ``medium" and ``hard" difficulty depending on the geodesic distance between start and goal locations.

\subsection{Path type curved, difficulty hard}
\begin{figure}[h!]
    \centering
    \includegraphics[width=1\linewidth]{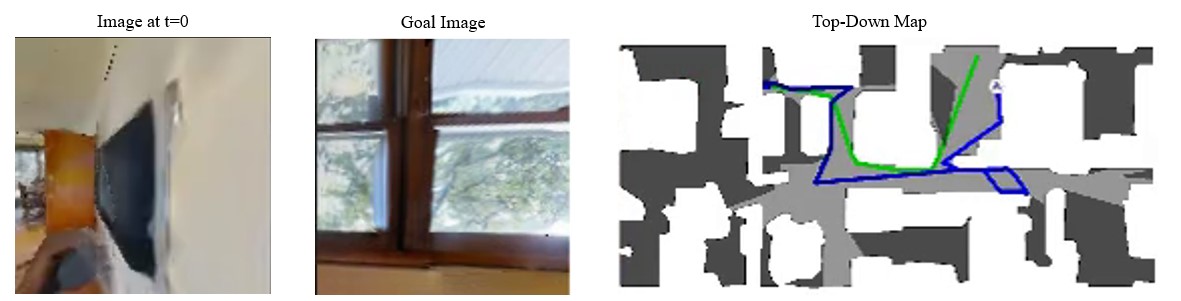}
    \caption{An expert trajectory on the ``curved - easy" split of the Gibson test set. The first image (left) shows the agent's observation at its start location. The second image (middle) shows the goal image, while the last image (right) shows the trajectory taken by the expert, indicated by the blue line. The green line indicates the shortest path between start and goal locations.}
    \label{fig:expert1}
\end{figure}

\begin{figure}[h!]
    \centering
    \includegraphics[width=1\linewidth]{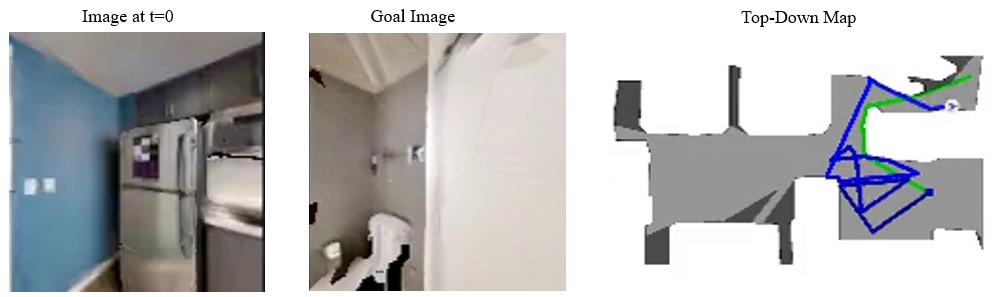}
    \caption{An expert trajectory on the ``curved - easy" split of the Gibson test set. The first image (left) shows the agent's observation at its start location. The second image (middle) shows the goal image, while the last image (right) shows the trajectory taken by the expert, indicated by the blue line. The green line indicates the shortest path between start and goal locations.}
    \label{fig:expert2}
\end{figure}

\clearpage

\subsection{Path type curved, difficulty medium}
\begin{figure}[h!]
    \centering
    \includegraphics[width=1\linewidth]{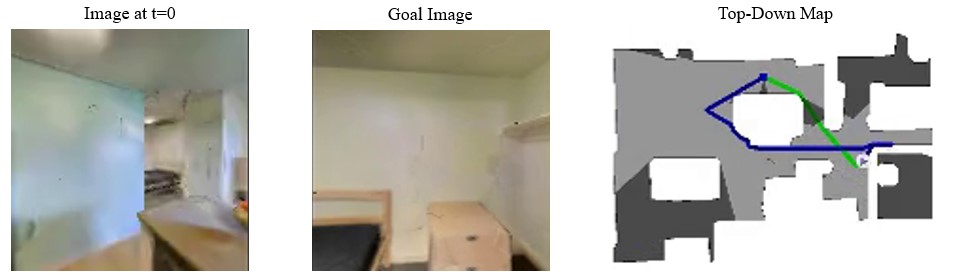}
    \caption{An expert trajectory on the ``curved - medium" split of the Gibson test set. The second image (middle) shows the goal image, while the last image (right) shows the trajectory taken by the expert, indicated by the blue line. The green line indicates the shortest path between start and goal locations.}
    \label{fig:expert3}
\end{figure}

\begin{figure}[h!]
    \centering
    \includegraphics[width=1\linewidth]{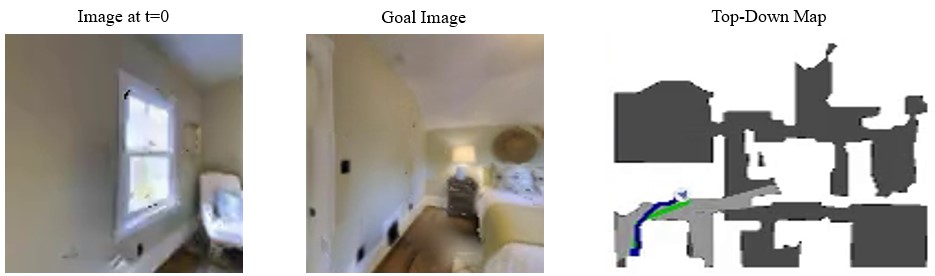}
    \caption{An expert trajectory on the ``curved - medium" split of the Gibson test set. The first image (left) shows the agent's observation at its start location. The second image (middle) shows the goal image, while the last image (right) shows the trajectory taken by the expert, indicated by the blue line. The green line indicates the shortest path between start and goal locations.}
    \label{fig:expert4}
\end{figure}

\end{document}